**Agent Teaming Situation Awareness (ATSA):**

**A Situation Awareness Framework for Human-AI Teaming**

Qi Gao[1], Wei Xu[2], Mowei Shen[1], and Zaifeng Gao[1]

[1] Department of Psychological and Behavioral Sciences, Zhejiang University

[2] Center for Psychological Sciences, Zhejiang University

**Author Note**



Correspondence concerning this article should be addressed to Zaifeng Gao,

Department of Psychological and Behavioral Sciences, Zhejiang University, 866 Yuhangtang

Road, Hangzhou, China. Email: zaifengg@zju.edu.cn



## Abstract

The rapid advancements in artificial intelligence (AI) have led to a growing trend of human-AI teaming (HAT) in various fields. As machines continue to evolve from mere automation to a state of autonomy, they are increasingly exhibiting unexpected behaviors and human-like cognitive/intelligent capabilities, including situation awareness (SA). This shift has the potential to enhance the performance of mixed human-AI teams over all-human teams, underscoring the need for a better understanding of the dynamic SA interactions between humans and machines. To this end, we provide a review of leading SA theoretical models and a new framework for SA in the HAT context based on the key features and processes of HAT. The Agent Teaming Situation Awareness (ATSA) framework unifies human and AI behavior, and involves bidirectional, and dynamic interaction. The framework is based on the individual and team SA models and elaborates on the cognitive mechanisms for modeling HAT. Similar perceptual cycles are adopted for the individual (including both human and AI) and the whole team, which is tailored to the unique requirements of the HAT context. ATSA emphasizes cohesive and effective HAT through structures and components, including teaming understanding, teaming control, and the world, as well as adhesive transactive part. We further propose several future research directions to expand on the distinctive contributions of ATSA and address the specific and pressing next steps.

## A Situation Awareness Framework for Human-AI Teaming

Situation awareness (SA), defined as the perception, comprehension, and projection of the environment elements within a volume of time and space (Endsley, 1988), has been identified as one of the key cognitive factors leading to safe and effective human-machine interaction, including air or road traffic domain (e.g. Gugerty & Tirre, 1996; Redding, 1992), and health care domain (Schulz et al., 2016), etc. As we continue to transit towards human interaction with AI systems, machines are evolving from mere automation to a state of autonomy, exhibiting unexpected machine behaviors and human-like cognitive/intelligent capabilities, including SA (Damacharla et al., 2018; Rahwan et al., 2019; Xu et al., 2022). When AI becomes a teammate of humans, the human-AI mixed team may gain potential performance over all-human teams, while SA adverse effects within the human-AI team have also been highlighted (McNeese, Demir, et al., 2021; McNeese, Schelble, et al., 2021). This distinction is primarily contingent upon the team process, thereby impelling us to delve into an investigation of the dynamic human-AI teaming (HAT). Considerable work has been done on how to achieve and maintain proper team SA within human-human teams, but the research focusing on HAT is still in its infancy.

Researchers have urged the field to proactively explore human cognitive mechanisms for modeling HAT (Liu et al., 2018; Xu et al., 2022), while the foundation of modeling SA remains limited. In this paper, a new framework for SA in HAT context is proposed based on the key features and processes of HAT. In the following sections, we will first introduce team cognition in HAT, and then elaborate on team SA. The third part of the paper is establishing the Agent



Teaming Situation Awareness (ATSA) framework. Finally, we discuss the application and summary of the framework.

## 1. Human-AI Teaming

The prevalence of HAT is irresistible, as it has been demonstrated to enhance performance in comparison to human-only or machine-only teams under various scenarios, especially in open-ended tasks with high uncertainties (Chen & Barnes, 2014; Cummings, 2014; Demir et al., 2017). Besides, AI possesses unique social advantages compared with a human as a teammate; for instance, AI can be designed not to judge. Therefore, assisted AI aiding special populations is in its prime, including autistic children (Jain et al., 2020), patients with movement obstacles (Budarick et al., 2020), and various professionals (Ziane et al., 2021), etc.

Human and AI can both be regarded as agents with intelligence and partial automation (J. D. Lee & Kirlik, 2013). Varying the behavioral pattern of AI can even blur the distinction from humans in a non-verbal Turing test (Ciardo et al., 2022). Therefore, AI has ushered in a new era of human-machine interaction, marked by a transition from human-automation interaction to human-autonomy interaction. The former is characterized by a subordinate relationship, while the latter resembles a teammate relationship. Automation fulfils its mandate in the confines of the program, regardless of the external context volatility, while autonomy is capable of analyzing information and making decisions adaptive to the situations through learning and generalization (Lyons et al., 2021; Vagia et al., 2016; Xu et al., 2022).

The autonomous characteristic of AI allows working together with humans as a team through Coordination, Cooperation, and Collaboration(3Cs, J. Lee et al., 2023). A "team" is made of "two or more individuals that adaptively and dynamically interact through specified



roles as they work towards shared and valued goals"(Salas et al., 2017). Distinguished from "groups", where members are largely independent and do not necessarily identify shared constructs, "teams" emphasize the interconnectedness of members. To achieve effective team control, team members have to arrange the timings of tasks and resources (coordination), use negotiation to resolve conflicts (cooperation), and make many decisions jointly over an extended time, developing shared rules, norms, and agreements (collaboration).

Effective team coordination, cooperation, and collaboration are supported by team cognition. Team members must harmonize cognitive and behavioral components to achieve their goal, which parallels the myriad of coordinated neuronal impulses that must coalesce to produce synchronized performance (Morrow & Fiore, 2013). Of all the cognitive components, SA is one of the most critical factors to be addressed in HAT (Chen & Barnes, 2014; Endsley, 2017), and is capable of integrating most relevant cognitive components through its process modeling, such as shared mental model and shared goal (Lyons et al., 2021).

However, so far, SA is only recognized as human metrics in SA-related models or HAT-related summaries (e.g., Barnes et al., 2019; Damacharla et al., 2018; Endsley, 2017), and machines are only external situation factors to be aware by humans. While the truth is that AI systems need to form SA as well, thus ensuring better coordination, collaboration, and adaptation abilities (National Academies of Sciences, Engineering, and Medicine, 2022). A remarkable SA framework for HAT should provide both the explanatory breadth and depth tapping into the HAT issues we reviewed here. To summarize, AI transacts its actions and cognitions, which embodies its autonomy, to form the team-wide control and understandings along with the human teammate counterparts.



Although some algorithm scientists have adopted the SA concept to improve AI design (e.g. Murray & Perera, 2021; Thombre et al., 2020), we argue that a top-down design process, directed by a HAT framework is necessary. While the SA process of AI shares some similarities with that of humans, its adaptability, programmatic nature, and lack of social interaction can all pose challenges for team SA in HAT, making it inappropriate to simply replicate human teaming strategies. To our knowledge, no such framework has been proposed that unifies human and machine behavior. As Woods (1998) stated, "Designs are hypotheses about how artifacts shape cognition and collaboration", a theory-derived SA framework accompanied by the elaborated cognitive process will fuel the hypotheses process, leading to efficient design. To develop a comprehensive framework facilitating better HAT, a review of leading SA theoretical models is performed in the following section.

## 2. Team Situation Awareness

A concise definition for SA is "knowing what is going on around" (Lundberg, 2015), while the definition enshrined in most articles on SA is "the perception of the elements in the environment within a volume of time and space, the comprehension of their meaning and the projection of their status in the near future"(Endsley, 1988). The latter is based on the most renowned theory of *Three-Level Model* of SA, which dominates the theoretical model, together with the Perceptual Cycle Mode (Smith & Hancock, 1995), to explain both individual and team SA. Although the two models serve to explain team SA from the individual decomposition perspective for a long time, a different system perspective, represented by the *Distributed SA* model (Stanton et al., 2006), ushered in a new era of the SA research, with the underlying theoretical emphasis shifting from the individual cognition to the joint cognitive system,



addressing the cognition embodied by both humans and artifacts. To fully understand the

decomposition of team SA, we hereinafter introduce first the two major theories describing

individual SA, and then theories for team SA from both individual and system perspective.

*2.1 Individual SA*

For the individual SA, two theoretical models, the Three-Level Model and the Perceptual

Cycle Model have survived in the ebb tide (Stanton et al., 2017; Wickens, 2008). In nature, the

two models apply different cognitive theories to explain SA. The three-level model takes

information processing theory as its foundation, while the perceptual cycle model directly

extends Neisser's perceptual cycle theory (Neisser, 1976).

As mentioned earlier, the three-level model forms the core by perception,

comprehension, and projection. Specifically, the concept lies in perceiving critical factors in the

environment (level 1), understanding these factors when integrating with goals (level 2), and

predicting the upcoming status change in the environment (level 3). However, the development

from low level to high level is not limited to a linear process, but rather includes the high level

directing the low-level SA. In this model, SA is the product of individual, environmental, and

system factors, consisting of a combination of pre-attentive processes, focused attention,

working memory, and long-term memory, coupled with alternating goal- and data-driven

processing (Endsley, 2021). The perceptual cycle model describes SA in a more macroscopic

manner, asserting that SA lies in the interaction between the individual and the external world.

The process of acquiring and maintaining SA is cyclical and encompasses three main phases

with schema, perceptual exploration, and environmental information. The internal mental

models (or schema) of current situations form anticipation of upcoming events, and directs the



course of perceptual exploration action, including checking whether the evolving environment

matches the mental model, which may require further search and explanation of the current

situation, thus modifying the existing model (Smith & Hancock, 1995).

Although intuitively one may feel that the two models differ starkly, Lundberg (2015)

pioneered a seminal integration of the two models. The three-level model depicts SA as the

core in a dynamic decision-making loop comprising the state of the environment, SA, decision,

and action execution (Endsley, 1995). If drawn in a circle, the similarity between the two

models appears (see Figure 1), as the SA section in the decision-making loop could be mapped

to the mental model section of the perceptual cycle. This view provides the field with a

promising direction to unify the SA models.

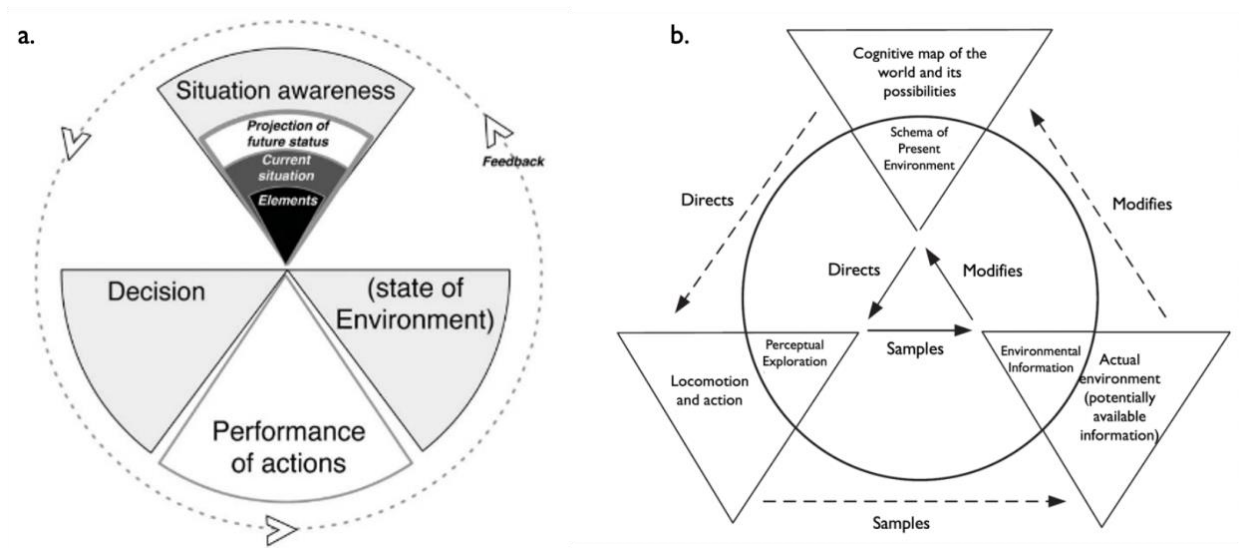

Figure 1. Two individual SA models, which have been redrawn to demonstrate
similarities between them. (a). Three-level model in dynamic decision-making (Endsley, 1995),
redrawn by Lundberg (2015). (b). Perceptual cycle model for SA (Smith & Hancock, 1995)
redrawn.

*2.2 Team SA*

As machines evolve into agents with intellectual ability, i.e., autonomies, they can be

deemed as humans' teammates (Tokadlı & Dorneich, 2022; Xu, 2020), thus extending the



connotation of SA in HAT to team elements, beyond the original environmental elements.  Each individual SA models had developed their team SA models separately. By Endsley's model, team SA involves all team-members knowing the relevant information for their individual role, and this includes individual SA only self-concerned and shared SA focusing on the sub-set of information needs that are in common(Endsley & Jones, 2001). Four factors need to be identified in support of good shared SA and team SA: 1) requirements, describing what information should be shared across team members; 2) devices, describing the media or the platform that sharing SA relies on; 3) mechanisms, describing the importance of the shared mental models; 4) processes, describing how effective team SA is achieved, for example clear delineation of tasks, goals, and roles. This theory primarily focuses on the breakdown of team SA outcomes, with little emphasis on the development and maintainence process of team SA.

Team process, while in the direct extension of perceptual cycle model of team SA proposed by Salas et al. (1995), is the core. In their model, team SA is resulted by the interaction between individual SA, including information processing functions and preexisting knowledge and dispositions, and various team processes, rather than simply addition for all SA that team members hold. The preexisting requirements for the teamwork, the characteristics of team members, and the team processes interact and affect one another. Therefore, individual and team sub-elements intertwine with each other resulting in team SA, and team SA in turn modifies these elements via team situation assessments process. A cyclical process of developing individual SA, sharing SA with other team members and then modifying both team and individual SA based on other team members' SA was addressed, which was inherited and adapted from the perceptual cycle theory.



However, we can see the commonalities between the two team SA models that they both dismantle team to only human team members, without taking artefacts like devices or automation machine into a whole SA system. Holding this systematic view, distributed SA argues that SA is held by and distributed between agents and artefacts comprising the system and can be viewed as an emergent property of collaborative systems (Salmon et al., 2017). Stanton et al. (2006) contended that each agent supplements other agents' SA with their own unique but compatible (not shared) views on the situation through interaction. Given that each agent possesses their own subgoals and subtasks, they do not require identical information to be shared, but rather only the information transacted within the team that can be 'decoded' by adjacent members. Accordingly, the SA partition exchanged among team members is referred to as 'transactive SA'. These two characteristics, namely compatible rather than shared SA and a systematic rather than cognitive view, comprise the key difference between distributed SA theory and previous individual decomposition team SA theories. We particulary value these notions in the HAT context because 1) the systemetic view is useful and suitable for AI team members with both artifact and human properties, and 2) compatible SA and transactive SA is more appropriate for complex, dynamic and occasionally ambiguous teaming processes.

**2.3 SA in HAT – human-human vs human-AI**

Due to the rapid development of technology, AI has become more capable as a teammate as opposed to simply a tool, highlighting the superiority of distributed SA theory. Many researchers have undertaken great efforts to tackle the SA problem in HAT, while most of them still adopt the individual cognitive perspective, which only considers SA of either human side or AI side and misses the other side. For example, the SA requirements for HAT raised by



National Academies of Sciences, Engineering, and Medicine (2022), focuses only on human side, encompassing situation, task environment, and teammate- and self-awareness. Besides, several other frameworks were raised as the design blueprint for AI. One of them is the cooperative intelligence framework integrating self + situational awareness for single vehicle and global SA provided by multi-vehicle cooperative sensing (Cheng et al., 2019). Another similar case covers cooperative perception (or sensing in the former model) and intention awareness (Kim et al., 2015). However, such attempts all eliminate the human role in non-fully autonomous AIs. Mostafa et al. (2014)  proposed a SA framework comprising human supervisor, system software and software agent. This framework endows system software with SA capability based on three-level model, and places the human and software agent as equal footing. Nevertheless, the resemblance between a human and AI is still not reflected, and no interaction between a human and AI is elaborated in this framework.

Chen et al. (2014) proposed the Situation Awareness-Based Agent Transparency (SAT) model to address interaction scenarios involving an AI. The SAT model extends three-level model by incorporating the 3P factors (Purpose, Process, and Performance, J. D. Lee & See, 2004) and the BDI agent framework (Beliefs, Desires, Intentions, Rao & Georgeff, 1995). The three levels of SAT model refer to: Level 1, goals and actions demonstrating what the agent is trying to achieve; Level 2, reasoning process that helps a human understand why the agent is doing so; Level 3, projections indicating what a human should expect to happen. In the original model, they only skimmed SA transparency on the operator side, while neglecting the AI side analogous to human SA. Therefore, the Dynamic SAT model was then developed encompassing the bidirectional, continuously iterating, and mutually interdependent interactions (Chen et al.,



2018). In the renewed model, system participants (including humans and AI) share the three levels to achieve the goals of their team tasks through feedforward and feedback loops. However, it fails to articulate that the foundation of transparency benefit in HAT is common activated mental models (Lyons, 2013). Besides, transparency is suggested not the only factor for successful HAT (Damacharla et al., 2018; Fischer, 2018; O'Neill et al., 2020).

Despite the emergence of other new SA concepts in HAT, such as supportive SA (Kridalukmana et al., 2020), mutual SA (X. Yuan et al., 2016), or cloud-enabled SA (Golestan et al., 2016), no systematic framework has been developed thus far. The overview highlights the urgent need for a model for SA in HAT that involves unified, bidirectional and dynamic interaction, ultimately promoting cohesive collaboration. To this end, based on the fundamental characteristics of HAT, we propose an Agent Teaming Situation Awareness (hereinafter referred to as ATSA) framework that builds upon previous SA models to guide effective HAT. It is worth noting that 'teaming', not 'team', is expressed in the continuous tense to underscore the dynamism and timeliness of the interaction and collaboration between human and AI.

### 3.  The Agent Teaming Situation Awareness Framework (ATSA)

We concur with the distributed SA theory that SA should be regarded as both a cognitive construct and a system construct. In the newly proposed ATSA framework (see figure 2), both human and AI cognitive constructs are represented by perceptual cycles as this intuitively reflects both the SA product and the process it inures (Stanton et al., 2017). Different from distributed SA theory, we employ cyclic representations as both the individual and team lens. We propose that the system construct of SA can also be represented by a similar



perceptual cycle as the cognitive construct: teaming understanding (TU, corresponding to 'mental model' in the individual cycle), teaming control (TC, corresponding to 'action' in the individual cycle), and world (all agents sharing the same external environment). Teaming collaboration is achieved from the iterative teaming cycles within a broader time scale (figure 3). Moreover, these two constructs in ATSA are linked through the transactive part, which is adapted from the 'transactive SA' concept in distributed SA theory. In this manner, the two constructs correspond with each other in a harmonious way.

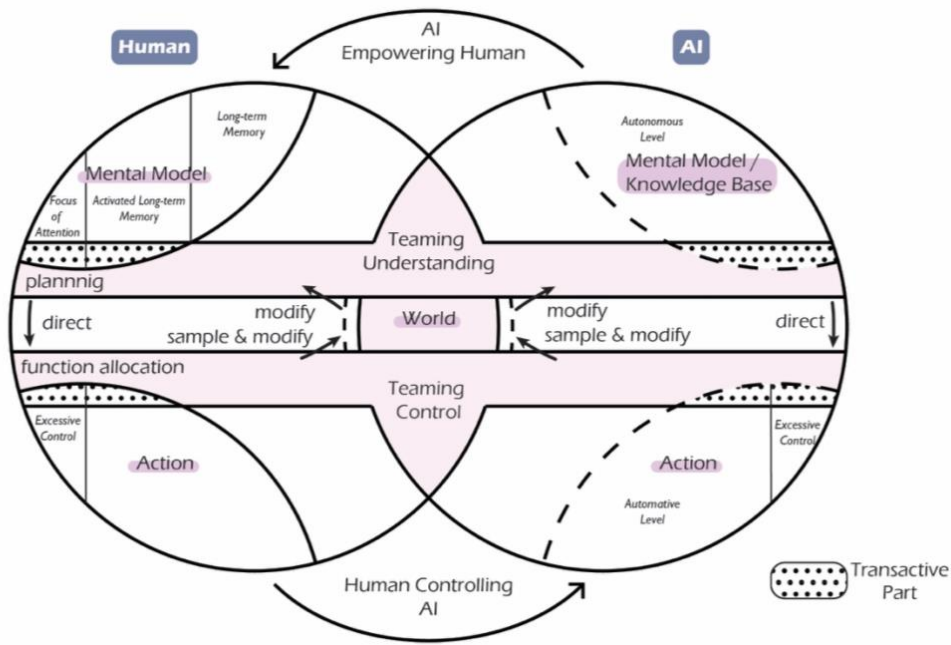

Figure 2. the Agent Teaming Situation Awareness Framework (ATSA). Two individual cycles (human cycle on the left and AI cycle on the right) and a teaming cycle (area with pink color fill) compose the overall frame.



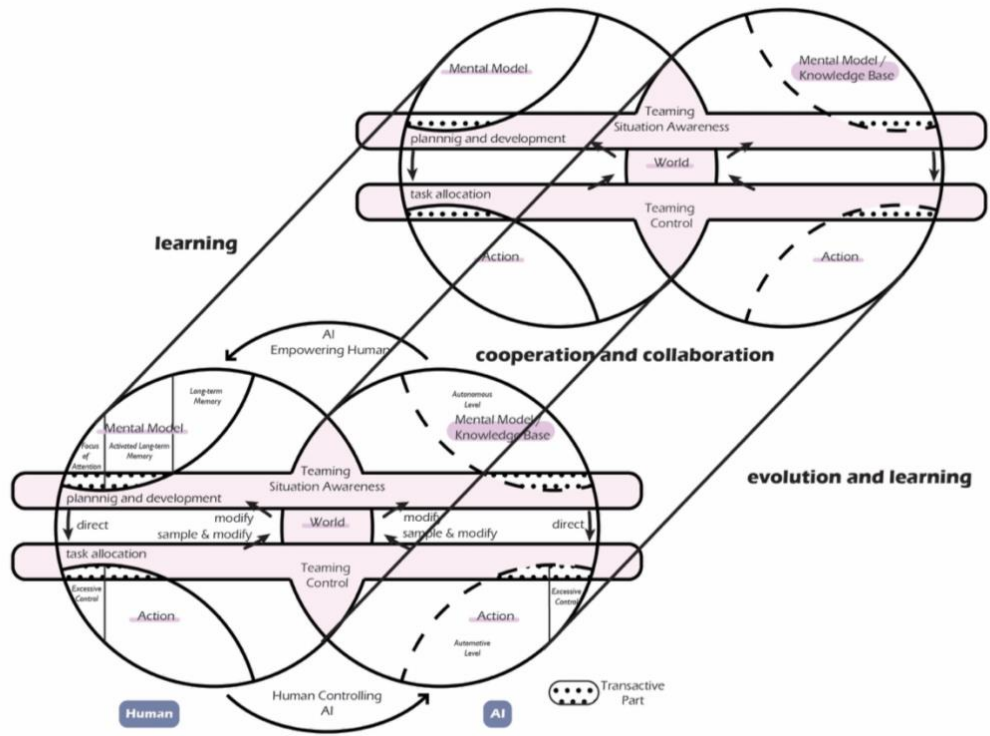

Figure3. ATSA expanded for teaming collaboration. Through the team co-learning and AI evolution, the team can manage better teaming collaboration as time progresses.

The subsequent section offers an extensive overview of the two crucial cycles within the ATSA framework for the individual construct and teaming construct.The individual cycle is a modified version of the original perceptual cycle theory, specifically tailored to cater to the unique requirements of the HAT context and cutting-edge cognitive findings. Here, we delve into the specific modifications made to the cycle and how they contribute to the overall effectiveness of the ATSA framework.The teaming cycle, on the other hand, has been emphasized and lucubrated on as it is the most direct part to ensure cohesive and effective HAT. We explore the structures and various components of the TU, TC and world in the teaming cycle and also elaborate on the adhesive transactive part. By providing an in-depth analysis of



both the individual and teaming cycles, this section aims to offer a profound understanding of the ATSA framework.

### 3.1 Individual Cycle

In each individual cycle, the agent, could be human or AI agent,  shares most of the world with their teammates, as the overlapped part demonstrated in the middle of the framework. The exploration of the external world modifies the mental model, which is transacted through a transactive part within the team, and formed into teaming understanding (hereafter referred to as TU). Ulteriorly, the planning is generated on the basis of TU, and is used to direct the function allocation. According to the function allocation, each individual agent acts for their individual task, through which achieving teaming control (hereafter referred to as TC). TC offers two routes for the team to interact with the world: sampling the world and modifying the world, thus closing the cycle.

Unlike the perceptual model modified by Smith & Hancock (2015) from Neisser (1976), we keep the link between action and the world. In Neisser's original model, the perceptual exploration serves as sampling the world, as well as modifying the world. Since this omitted control process contributes to the modification of the internal mental model, and the control process is another critical issue for HAT (e.g. Huang et al., 2022; Marcano et al., 2020), we revert this path for SA in our framework. The necessity for the incorporation of control process and the cognition process is supported by extended control model, which builds on the perceptual cycle model just as the extended SA model (Hollnagel & Woods, 2005).

Different from the original perceptual cycle model, we herein do not perpetuate the phenotype and genotype decomposition of mental model in humans, rather postulate that



individual SA product is embodied in the activated mental model, with higher information priority and availability than other information in the mind. With activated mental model, we extend the original mental model, which refers to the abstract long-term knowledge structures humans employ to describe, explain, and predict the world(Johnson-Laird, 1983), to working memory structure. SA was believed to be reside in working memory for a long time(Johannsdottir & Herdman, 2010), while recent studies have clarified that SA relies on a more integrated interaction between the working memory and long-term memory system, rather than only on working memory (Endsley, 2015; Wickens, 2015). This accords with cognitive research on working memory, which has increasingly been referred to as activated long-term memory (Cowan, 1988). In this view, the ensemble of components in working memory are in a heightened state of availability for use in ongoing information processing (Logie et al., 2020) rather than being a separate structure from long-term memory (Baddeley, 2012). We parallel the activated mental model concept with working memory as the activated long-term memory, and propose that only the active part of the generalized mental model interacts with TU directly.

As for the cycle of AI, several different aspects have been identified from that for human. Firstly, since both software and hardware are designed by humans, and the flexibility exists even after the AI is in operation(e.g. Akula et al., 2022; Li, 2017; L. Yuan et al., 2022), we adopt dashed lines for mental model and action of the AI to signify the adjustability of these two parts. Besides, the area of each rugby shape represents the corresponding capability. For AI, the division of mental model and action ability highlights the difference between autonomy level and automation level. "Automation" focuses on the extent to which machines can



contribute to a task, emphasizing on the acting ability. On the other hand, under different automation conditions, tasks that would otherwise be executed by humans are partially or completely delegated to machines. In contrast to automation, "autonomy" highlights the independence and dynamism of machines to complete the task, even when the solution was not programmed, relying on the human-like information analysis ability (Endsley, 2017; Simmler & Frischknecht, 2021; Xu, 2020).

Though the viewpoint of autonomy analogues to a human is stressed, ATSA is a human-centered framework in nature, which is embodied in two ways, corresponding to TU and TC respectively (Xu, 2020; Shneiderman, 2020a). First of all, plastic AI are designed to empower human, rather than replace human. AI should proactively align their mental model to human (value alignment, L. Yuan et al., 2022) and form TU. This is in line with Shneiderman's view except he describes diametrically from another extreme that portrayals of tool-like appliances for AI is unacceptable (Shneiderman, 2020a, 2020b). Secondly, ensure human ultimate control. This requires the AI systems keep humans access to the final decision, explain enough for the action they take, and provide humans with meaningful control (Beckers et al., 2019; de Sio et al., 2022). Notably,  two similar facets were summarized in Rieth and Hagemann (2022), by stressing considering human needs and human-strength based function allocation, which can also be derived from TU and TC components.

### 3.2 Teaming Cycle

#### 3.2.1 Teaming Understanding

TU is the mental product of the team SA. In ATSA, TU is deconstructed into three orthogonal dimensions (see Figure 4): content, process, and state. Content, explaining what is



in TU, pertains to the outcome manifestation arising from the amalgamation of SA elements. While process, elucidating the temporal sequencing of contents, provides insights into the development stage of the product. Finally, state, capturing how the contents are represented, encapsulates the priority assigned to the products in mind.

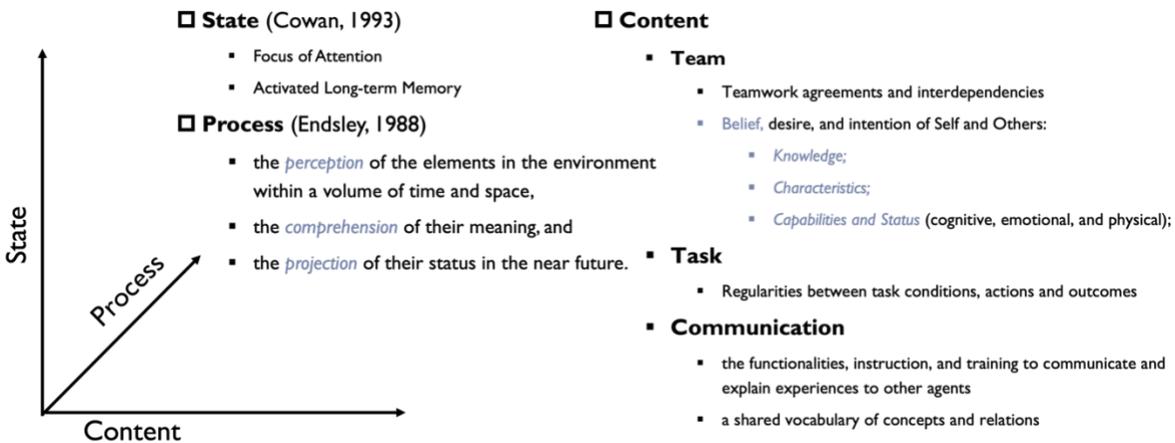

Figure. 4 Three dimensions of TU: State, process, and content.

Though some researchers believe the elements of SA are highly domain-specific (M. R. Endsley, 2021), we posit that there exist common categories that undergird the HAT process: team, task, and communication, addressing who, what and how respectively. Whereas a conventional human-human team necessitates each team member to possess a task model and team model (Converse et al., 1993; Scheutz et al., 2017), in HAT, the communication model assumes particular significance, given that the interaction between humans and AI may not be as natural as human-human or even AI-AI. The shared mental model and shared SA field have both sought to encapsulate the content that HAT requires, and we argue that information from both working memory and long-term memory is interwoven, thereby rendering it unsuitable to disentangle the two sources. Van den Bosch et al. (2019) delineate six mental model challenges for human-AI co-learning, and National Academies of Sciences, Engineering, and Medicine



(2021) put forward that models for teammate (including self) and model of the world are necessary. We modified and integrated these models into the aforementioned three common categories so that they fit in a framework cohesively. To form TU, or the so-called value alignment process (L. Yuan et al., 2022), the human-AI team must arrive at a consensus on these three components.

Regarding the *team* component, establishing a shared set of teamwork agreements and interdependencies is of paramount consideration, for example, agreements on how agents dynamically update these agreements as continuous interaction. Besides, an essential prerequisite for effective team performance is the awareness of self and others in the team (Andrews et al., 2022; Sycara & Sukthankar, 2006), including the belief, desire, and intention (BDI (Rao & Georgeff, 1995). Moreover, belief comprises knowledge, characteristics, capabilities and status. Capabilities encompass cognition (e.g. memory, attention, etc.), emotion (e.g. empathy, emotion intelligence, etc.) and physical capability (e.g. action limits, etc.). Status includes cognitive resources (e.g. vigilance, engagement, etc.), emotion states (e.g. happy, nervous, fearful, etc.), and physical availabilities (e.g. drunk, hurt, etc.). As for the *task* component, the team should be aware of regularities between task conditions, actions, and outcomes, as this enables planning based on rational reasoning. Moreover, the current task content and condition should be agreed upon within TU, which is decisive of the current frame of interaction. What is important for the next frame of interaction is that the prediction of the outcome, which also requires in common understanding. Finally, the *communication* component is vital to the team process that culminates TU. This encompasses not only a shared



vocabulary of concepts and relations, but also the functionalities, instructions and training to communicate and explain experiences to other agents.

The second dimension of TU is process, which describes the information analysis process from a cognitive perspective. Endsley's three level model can be adopted to address this dimension. Though there are some misconceptions claiming that the three levels are strictly linear (e.g. Salmon et al., 2012; Sorensen et al., 2011), Endsley has clarified the SA process as an ascending but non-linear development path (Endsley, 2015).

The third dimension of TU is the information state. As previously mentioned, there's no strict distinction between information in working memory and long-term memory; it is merely a matter of priority. Information in activated long-term memory has a higher access priority than that in long-term memory, while critical role of attention and control in shuttling items into and out of the focus of attention (FOA), allowing information has higher priority than that in activated long-term memory (Cowan, 1988). For example, whereas activated memory items, or items in FOA, function as an attentional template and directly affect perception, accessory items in activated long-term memory do not (Olivers,2011; Oberauer, 2012). Therefore, an additional FOA is included in the mental model to represent the information with highest information priority.

**3.2.2 Teaming Control**

TC is the behavioral achievement of the team SA. In ATSA, two hierarchical dimensions compose TC: complementing flow and function allocation, addressing *how and what* to control respectively. Complementing flow depicts the relationship between the agents, determining



the assignment of task control authority, which is further reflected in the allocation of function and tasks.

The complementing flow describes the top-level design for maximizing hybrid intelligence for HAT, which is based on the synthesis analysis of constraints, costs, quality, and availability of human engagement (Kamar, 2016). This hybrid intelligence must be propelled by leveraging the complementary advantages of AI and human intelligence (Dellermann et al., 2019). In the team in which they try to compensate for their weaknesses, it might be human towards AI or AI towards human, or peer to peer complementing where both agents complement each other (Zahedi & Kambhampati, 2021). When AI is complementing humans, complicated human cognitive dynamics are involved, introducing AI interpretability need to meet human motivations, expectations, or calibrate trust. When humans complement AI , they usually monitor the AI to prevent mistakes, failures, and limitations, thus improving the AI system's performance. Under this complementing flow, human interpretability, such as behavioral and physiological metrics indicating their status, becomes vital. For example, out-of-the-loop problem could bring huge damage in some life critical systems, such as high-level autonomous driving (Endsley, 2021). For a peer-to-peer teaming, AI and humans may interchangeably enter the land of one another for action, decision, or coordination. The bidirectional communication and feedback between the two entities will help achieve a more effective teaming. Though these three kinds of complementing flow diverge in some points, they are all closely related with how to achieve TU, and how to convey TU across team members.



Moreover, note that "human complements AI" does not conflict with human centered AI, as the latter emphasizes that the ultimate decision authority belongs to human, and the AI should defer the control authority to human ultimate decision authority. Control authority can be transitioned when needed, for example, when human monitors the AI malfunction for risk management (Bellet et al., 2019), which is well-studied in autonomous vehicle takeover issues (Maggi et al., 2022). This human-centered view is in line with the human-directed execution principle proposed by Battiste et al. (2018).

Complementing flow is further embodied and crystallized by function allocation, regarding how system functions and tasks should be distributed across humans and AI (Roth et al., 2019). Static function allocation was prevailed in early days, when the automated function must not be too critical, or too central for human activities. These traditional approaches, for example "Men-are-better-at/Machines-are-better-at" (MABA-MABA, Fitts, 1951) or Level of Automation (LOA) framework (Parasuraman et al., 2000), list pre-determined strength-based human and machine work (Rieth & Hagemann, 2022), yet without considering dynamically changed contexts and finer-grained needs (Jamieson & Skraaning, 2018; Johnson et al., 2018; Roth et al., 2019), thus underscoring the necessity of dynamic function allocation. Dynamic function allocation, or adaptive automation, proposed to a shift from a strict LOA perspective to a cooperation modes perspective. Hoc (2013) asserted that HAT can fall into five modes in general: (1) Perception mode: The AI is designed only in order to enhance the human perception; (2) Mutual control mode: the AI criticizes human behavior in relation to a standard; (3) Shared control mode: The agents act at the same time on the same variable; (4) Function delegation mode: one of the task functions instead of all the functions remaining under the



human control; (5) Full automation: humans do not need to interfere at all. Throughout the interaction, the function allocation can shift among these five modes. Moreover, action complementary mechanism should be designed in human-AI systems to prevent excessive control (e.g., suicidal pilot, drunk driver) via the combination of mutual control mode, shared control mode, and function delegation mode.

### 3.2.3 World

World is both the constraint and the behavioral product of the team SA, the former of which is the start point of SA, and the latter is the end point of SA. Less is directly discussed about external world regarding HAT, yet we've seen its presence in almost all SA or control models through sensation input or feedback (e.g. Salas et al., 1995; Woods & Hollnagel, 2006). Both constraints and behavioral products include various non-adaptive environmental context (natural, social, and technical), and adaptive system devices (hardware, software, and interface).

Note that there is a non-overlapped area for each individual agent indicated by a dash line outside the teaming world part. This area demonstrates the world sampled or modified by this single agent, which can, at least partially, reflect the "control level" of this agent as expounded by Shneiderman (2020a, 2020b). In his two-dimensional human-centered AI framework, the human and AI control level are two orthogonal dimensions, suggesting even highly autonomous system could still keep people in high control level.

## 3.4 Transactive Part

Transactive part is the aggregation of elements transacted between team part and individual part, which ensures that SA is both teamed and individual-owned. Therefore, the



interaction between the individual part and the teaming part is accomplished by transaction process (a kind of team process) and the transactive part. Access to information relies on knowing who knows what, or transactive memory (Wegner, 1987). Transactive SA was first proposed by Stanton and Salmon et al. in the distributed SA theory (Salmon et al., 2017; Stanton et al., 2006, 2017). Via transactions in awareness that arise from sharing of information, distributed SA is acquired and maintained. We draw on this line and postulate that TU and individual mental model shall communicate, interact and transact to each other through the transactive part, where team process can exert its influence on. The integration of the transactive part forms the teaming part, and the teaming part is what is relevant to achieving team goals, while the individual part may encompass some non-team-goal-relevant elements. The same is true for TC and individual action. Excessive individual control should be regulated by the interlock of individual transactive part, hence maintaining TC at an acceptable level.

One of the obstacle of the transaction is a phenomenon that Steiner (1972) described as "process loss"—a decrement in coordination that results in performance below team potential. While initially discussed in the context of human teams, this concept can be applied equally in HAT context. There have been several studies (e.g. You et al., 2022; Morrow & Fiore, 2013; Baker et al., 2019; Schneider et al., 2021) addressing the problem through designs and methods that mitigate process loss in HAT via devices in the *world*. Cognitive Interfaces are extensively studied to enhance human cognition of AI teammate, for example transparency-related studies (You et al., 2022). Morrow and Fiore (2013) elaborated upon the value of external representations in the world, such as process mapping, which provides a visual representation of work flow. Collaborative problem conceptualization often requires reshaping, or even



discarding the puzzle pieces from the information in the process map. Negotiating the construction of the map itself can lead to the development of TU. Moreover, the Real-time Flow, Event, and Coordination Tool (REFLECT, Baker et al., 2019) has been developed to capture data for mapping communication flows within the team, including who speaks to whom throughout team interactions. However, the effort is limited in its ability to capture the full range of team processes, including implicit and explicit communication, negotiable and directive requests, and other factors (Schneider et al., 2021).

### 3.5 Teaming Cooperation and Collaboration as Temporal Extension

According to J.Lee et al.(2023), the three teaming forms (coordination, cooperation, and collaboration) are connected by adaptive cycles of different time scales and resiliences. A single ATSA frame reflets coordination, which is brittle, meaning that it is vulnerable to unexpected threats and has low adative capability. While multiple ATSA frames iterations constitute cooperation and collaboration (see figure 3), with higher resilience and longer time constants. Collaboration is often used interchangeably with cooperation, but collaboration is a higher resilient and longer co-learning and evolving process in which teammates share and update long-term knowledge, rules, and goals to make decisions jointly (McNamara, 2012). In collaboration, the complementing flow becomes peer-to-peer teaming, with AI requiring little supervision (Mainprice et al., 2014) and engaging in active coordination with human peers to exchange ideas, resolve differences, execute tasks, and achieve goals (Rule & Forlizzi, 2012). Attitudes (e.g. trust, disposition to collaborate) are developed along the cooperation and collaboration process. Compared to cooperation, collaboration can be viewed as a long-term relationship between human and AI, as most aspects of collaboration are uncertain, while



cooperation is determined by their fixed shared values and goals. During the cultivation of this long-term relationship, both TU and TC, and therefore individual mental models and actions, will dynamically change over time.

### 3.6 Discussion

Overall, ATSA is a SA framework for HAT context. It takes into account the unique characteristics of both human-human teams and human-AI teams, while also recognizing the similarities between humans and AI as autonomous and intelligent agents. In Table 1, extant theories that can be used to elaborate SA in HAT are compared.

Table 1. Comparison between Team SA, Distributed SA and ATSA

|  | Team SA by Endsley | Distributed SA | ATSA |
|---|---|---|---|
| Target Teams | All human teams; intelligent machines are devices required for team SA | Hybrid teams or all-human or all-machine teams; it is especially designed for complex systems | HAT, with different characteristics described for human and AI; while either half of the framework can be substituted to accommodate all-human or all-machine teams |
| Nature of SA | Cognitive construct | Both cognitive construct and system construct | Both cognitive construct and system construct |



| Approach of Forming Team/Teaming/Distributed SA | Individuals share their own SA, thus forming shared SA with same contents | Distributed SA is a characteristic of the system, independent of team member entities; individual SA are compatible and can be transacted | Team SA is embodied in the teaming cycle of TU, TC, and the world; individuals interact with the teaming part through transactive part |
|---|---|---|---|
| Theoretical Foundation | Three-level model of Endsley | Joint cognitive theory and perceptual cycle model | Joint cognitive theory and perceptual cycle model |

ATSA provides a solid theoretical basis for designing HAT system. The peculiar insights it provides us are elaborated as follows.

(1) Three end components

First, ATSA divides the entire HAT succinctly into three components, *Teaming Understanding, World, and Teaming Control*, covering all aspects of the interaction process. What really matters are TU and TC. No matter how the process changes or deteriorates, team performance will not be influenced as long as TU and TC maintain well.

(2) Different priority for SA

ATSA separates the information priority status of information in individual mental model and TU, with different priorities contributing differently to team



performance, making it necessary to distinguish the key SA components and ensure them appropriate priorities under particular scenarios. Meanwhile, there is a significant difference between AI and humans regarding information priority, since AI can treat all information equally or adjust priorities more flexibly, while humans are restricted by cognitive limits (e.g. Gao et al., 2022). Therefore, the interaction entailed by this difference is worth exploring.

(3) Transactive part matters

Third, since *mental model* and *action* are both passed through the transactive part and integrated into the teaming part, and the team work together through the teaming part to complete the plan and function allocation, the most critical factor for team process is the transactive part rather than individual part.

(4) Evolutionary AI

ATSA corresponds AI automation level and autonomy level to *Action* capability and *Mental Model* capability respectively, which are more plastic than humans. In this way, HAT research for function allocation may be clearer when tapping to questions related to automation and autonomy. Besides, different from human innate fixed cognitive and physical mechanism (e.g., selective attention), AI algorithm which decides cognitive mechanism can evolve actively through iteration (e.g. Gupta et al., 2021), and both hardware and software can be adjusted passively by the engineer. Therefore, what will the flexibility of AI teammate bring to team dynamics remains an open question.

(5) Teaming collaboration dynamics



ATSA effectively addresses cooperation and collaboration issues by extending the

time horizon. As AI evolves and teams engage in co-learning, both TU and TC are

able to continuously update, leading to the emergence of novel and complex

teaming dynamics, such as attitudes, strategies, and long-term relationships, which

play a pivotal role in fostering effective and cohesive collaboration.

Though ATSA is capable of tackling many issues in HAT, there may remain some facets

to be improved. A critic of this model might be that ATSA apply a single human and AI

perspective (Renner & Johansson, 2006; Salmon et al., 2018). We are gratifying to note that

ATSA is endowed with good extensibility, as we may add as much agents as we want if we

convert 2D view to 3D view (see Figure 5). This allows ATSA to be accountable for complex

systems with multiple humans or AI, while retaining its elegant essence. Another criticism might

be that ATSA is unable to interpret issues like trust (e.g. Biondi et al., 2019; de Visser et al.,

2018; Rebensky et al., 2021; Xing et al., 2021). Though ATSA is a cognitive model in nature, we

argue that social factors like trust are able to act upon the cognition process (i.e. social

cognition) and the cooperation/collaboration process. In this way, trust can not only influence

the TU, but also the consequent TC (e.g. Adams et al., 2003; Dzindolet et al., 2003). Therefore,

social dynamics may be extended to the model in the future.



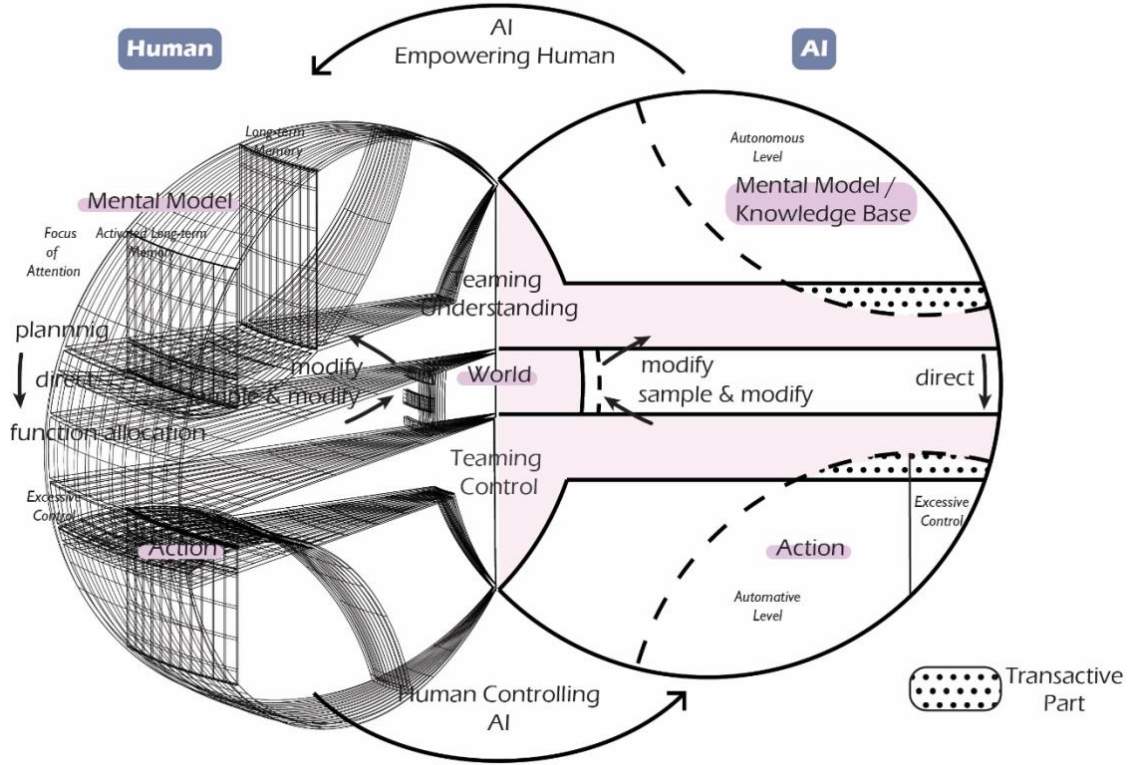

Figure 5. ATSA 3D conversion. Practitioners can insert as many agents as necessary.

## 4.  Future Work and Application Case of ATSA

### *4.1 Next Steps*

#### 4.1.1 Teaming SA Measurements

We posit that teaming SA is a system-level emergent property that cannot be segregated into individual components and measured separately. Prior research has attempted to do so by taking into account the individual SA additions as the team SA result (e.g., Gombolay et al., 2017). However, we emphasize the dynamic nature of SA in ATSA. Consequently, a final test after the task's completion is not a reliable measure of teaming SA. Instead, we recommend a direct in-process approach, which requires urgent attention so that ATSA-related empirical research can be conducted.



Prior research on direct in-process measurement for team SA has provided some insights into measuring teaming SA. Two approaches that may be useful.

(1). Team probe-recall techniques (Bolstad et al., 2005) involve using a Situation Awareness Global Assessment Technique (SAGAT) in a team environment, where the thask is frozen and participants have to recall information related to perception, comprehension, and prediction. Although this approach only captures a small part of teaming SA (the process of TU), we suggest extending checklists for information matrices within the three dimensions (contents, processes, states) into the probe-recall as a comprehensive measurement for teaming SA. Furthermore, real-time neural decoding of individual focus of attention (FOA) may also be useful to reveal agreements or conflicts of TU.

(2). The Co-ordinated Assessment of Situation Awareness of Teams (CAST, Gorman, 2006) employs a scenario involving "roadblocks" and assesses team SA by evaluating how the team perceives and responds to these obstacles through coordinated actions, resulting in a measure of team SA. This introduces a process-performance-based team SA measurement that synthesizes both cognition and action, which approximates ATSA to a large extent. However, it still fails to measure the concrete dynamic happening within the coordination cycles. A better solution might be to establish TU and TC networks according to the roadblock task settings and evaluate the network correspondingly.

### 4.1.2 Proper Teaming SA within Cooperation and Collaboration Dynamics

Most SA related studies focus on enhancing team SA, with the assumption that higher SA leads to improved team performance. This belief is prevalent among researchers, with some even claiming that SA can be indirectly measured through performance (Endsley, 2021).



However, the relationship between SA and team performance may not be strictly positive, particularly in the context of teaming SA in HAT. We argue that only critical information relevant to the current task in FOA and activated long-term memory, as well as prediction information related to subsequent tasks in activated long-term memory but not FOA, are sufficient and necessary for achieving higher team performance.

For the HAT context, positive relationship might be right for systemetic level teaming SA, while not for cognitive level individual SA. Peer-to-peer teaming requires both higher level teaming SA and individual SA, whereas non peer-to-peer teaming might not require this level, as it involves the consideration of various factors including costs, constraints, quality and availability (Zahedi & Kambhampati, 2021). When the AI complements the human or vice versa, teaming SA requirements remain the same as peer-to-peer teaming, while the subordinate member may not require as much individual SA as before. Instead, they must find a balance among multiple factors such as workload, time cost, and attention. Moreover, when more complex factors such as trust and team harmony are introduced in cooperation and collaboration, both teaming and individual SA requirements may change, at least on the 'team' component. Therefore, it becomes crucial to explore in depth how teaming SA can be calibrated in the HAT context.

### 4.2 Usage of ATSA in an Autonomous Vehicle Case

Take HAT in an autonomous driving accident as an example, where the framework should be applied in all changed time points, including normal ADAS mode, hazard appearing, warning, human takeover, and after human takeover. When the AI, for instance, adaptive cruise control system, is controlling the vehicle, and the human is immersing in a non-driving



related task, for example watching movies, TC is under function delegation mode. Though both

driver and AI reach TU on team and task elements, they cannot reach agreement on

communication element if AI does not aware that the human engagement of visual and

auditory modality. Once the AI notice the hazard and try to warn the driver through visual and

auditory signals, what it does is transacting its activated mental model to the driver. However,

the driving-related information priority for the driver now is much lower than the contents in

the movie, and plus the cognitive limits restrict him/her capture the signal (Inattentional

blindness, Mack, 2003). When the driver is finally aware of the AI warning, the AI may have

already developed the TU to the comprehension or even prediction level, and this TU transacts

back to the driver through multiple kinds of devices in car, such as head-up displays, or voice

assistants. The driver successfully takes over the control, while TC remains shared control mode

in that AI and the driver is controlling some specific parameters (e.g., handwheel) together, as

humans need a recovery phase when getting back in the loop (e.g., Russell et al., 2016). As both

the human and the AI confirm the hazard has been successfully passed, i.e., they reach TU on

the task elements, the TC mode shift back to function delegation mode.

Through this analysis, we can easily identify several important issues discussed in the

HAT field, such as transparency, intention, and workload etc. (e.g. Biondi et al., 2019; O'Neill et

al., 2020). By ATSA, transparency is not only a design principle for the adaptive system devices

or displays in the *World* part, but also a principle for mental model and action transaction for all

agents. The reason why intention becomes a research hotspot is that it ties TU and TC most

closely. As to workload, together with intention, they are one of the most rapidly changed



components during the HAT process, compared with other SA components, such as task or communication.

Moreover, ATSA can be practically applied to guide engineering implementation. Here we propose several implementation directions that leverage ATSA's unique capabilities. Overall, ATSA is able to guide an interaction or collaboration work flow for hybrid human-AI team, albeit in a highly context-specific manner. Firstly, for instance, an ATSA-based algorithm concept can serve as a threshold to facilitate dynamic function allocation between the human and the autonomy in terms of mental model and action level of both entities. Second, algorithms can be designed to enhance teaming SA by computing TU and TC conflict levels and thus resolving corresponding conflicts.

## 5.  Concluding Comments

In summary, we conducted a comprehensive review of SA-related frameworks in HAT, and conceptualized a novel, multi-directional, dynamic and theory-based framework (i.e., ATSA) for human-AI interaction and collaboration. ATSA is a cyclical SA process which contains three interrelated components: world, teaming understanding, and teaming control. Despite the dissimilarities between human and AI, both parties transact their individual mental models and actions to form teaming understanding and teaming control, which in turn affect and modify the individual aspects. We anticipate that ATSA will contribute to both the theoretical foundations and application domain in HAT.

We propose several future research directions to expand on the distinctive contributions of ATSA and address the specific and pressing next steps. Firstly, it is necessary to conduct empirical research to validate the effectiveness of ATSA and its unique contributions in



real-world HAT, particularly regarding the different priority of SA and the transactive part. This could involve conducting experiments or simulations to assess the impact of different levels of TU and TC on team performance and exploring how different SA components contribute to team performance under various scenarios. Analytic network process tools might help in solving such scenarios. The premise of these empirical research is to establish direct in-process approaches to measure teaming SA, which requires partical and convenient approach to produce satisfactory measurement results. Lastly, it is crucial to explore the optimal level of teaming SA in HAT, and how it can be calibrated in the context of cooperation and collaboration dynamics. This could involve developing new benchmarks for evaluating teaming SA in different contexts with mathematic approaches. Addressing these research directions will help advance our understanding of ATSA and therefore HAT systems, leading to more effective and cohesive collaboration between humans and AI.